\documentclass[letterpaper, 10 pt, conference]{ieeeconf}  

\IEEEoverridecommandlockouts                              

\usepackage{graphicx}
\usepackage{booktabs}
\usepackage{amsfonts}
\usepackage{amsmath}

\usepackage{bbm}
\usepackage{algorithm}
\usepackage{algpseudocode}

\usepackage{multirow}
\usepackage{amssymb}
\usepackage{bbding}
\usepackage{soul, color}
\setstcolor{yellow}
\soulregister\cite7
\soulregister\ref7
\usepackage{listings}
\usepackage{xcolor}
\usepackage{url}
\usepackage{bm}
\usepackage{hyperref}

\title{\LARGE \bf
TLA: Tactile-Language-Action Model for Contact-Rich Manipulation
}

\author{Peng Hao$^{1}$\textsuperscript{*}, Chaofan Zhang$^{2}$\textsuperscript{*}, Dingzhe Li$^{1}$, Xiaoge Cao$^{2}$, Xiaoshuai Hao$^{3}$, Shaowei Cui$^{2}$, and Shuo Wang$^{2}$
\thanks{* These two authors contribute equally to this work.}
\thanks{This work was supported in part by the National Key Research and Development Program of China under Grant 2023YFB4705000, in part by the National Natural Science Foundation of China under 62303455, 62373039, U23B2038, 62273342, and 62122087, in part by Beijing Natural Science Foundation under Grant L233006, in part by the Youth Program and the Open Projects Program of State Key Laboratory of Multimodal Artificial Intelligence Systems. (Corresponding author: Shaowei Cui (shaowei.cui@ia.ac.cn))}
\thanks{$^{1}$Samsung Research China - Beijing (SRC-B), Beijing 100028, China}
\thanks{$^{2}$The State Key Laboratory of Multimodal Artificial Intelligence Systems, Institute of Automation, Chinese Academy of Sciences, Beijing 100190, China}
\thanks{$^{3}$Beijing Academy of Artificial Intelligence, Beijing 100083, China}
}
\begin{document}
\bstctlcite{References:BSTcontrol}

\maketitle
\thispagestyle{empty}
\pagestyle{empty}

\begin{abstract}
Significant progress has been made in vision-language models. However, language-conditioned robotic manipulation for contact-rich tasks remains underexplored, particularly in terms of tactile sensing. To address this gap, we introduce the Tactile-Language-Action (TLA) model, which effectively processes sequential tactile feedback via cross-modal language grounding to enable robust policy generation in contact-intensive scenarios. In addition, we construct a comprehensive dataset that contains 24k pairs of tactile action instruction data, customized for fingertip peg-in-hole assembly, providing essential resources for TLA training and evaluation. Our results show that TLA significantly outperforms traditional imitation learning methods (e.g., diffusion policy) in terms of effective action generation and action accuracy, while demonstrating strong generalization capabilities by achieving over 85\% success rate on previously unseen assembly clearances and peg shapes. We publicly release all data and code in the hope of advancing research in language-conditioned tactile manipulation skill learning. 
Project website: \url{https://sites.google.com/view/tactile-language-action/}.

\end{abstract}

\section{INTRODUCTION}
Tactile perception is crucial for contact-rich robotic manipulation tasks~\cite{billard2019trends}. For example, in fine assembly tasks, robots need to precisely sense small variations in the surface of an object~\cite{cui2021hand,cui2021toward}. Tactile sensing allows robots to make subtle adjustments in their contact pose, avoiding damage or misalignment~\cite{calandra2018more}. This precise perception of contact is indispensable in many complex tasks~\cite{li2020review}. Previous studies have shown that the integration of tactile feedback significantly enhances the flexibility and robustness of manipulation skills~\cite{wang2020swingbot,qi2023general}, especially when dealing with complex or unforeseen contact environments, thereby improving the robot’s adaptability. However, current methods largely rely on specialized models trained on specific datasets~\cite{lee2020making,dong2021tactile}, which are limited in terms of generalization and cannot match the capabilities of general-purpose models~\cite{team2024octo,brohan2023rt}.

Recently, large language models have made significant breakthroughs in human-like reasoning~\cite{naveed2023comprehensive}, with rapid progress in the development of vision-language-action (VLA) models~\cite{kim2024openvla,wen2024tinyvla}. By cross-modal language grounding, VLA models outperform traditional imitation learning methods, particularly in terms of generalization across different robot platforms and task settings. However, most current VLA models are primarily focused on visual tasks~\cite{o2024open}, lacking crucial tactile modalities, which limits their applicability in contact-rich manipulation tasks~\cite{zhou2023language}.

Despite recent efforts to align language and tactile for perceptual tasks~\cite{yang2024binding,cheng2024touch100k,fu2024touch}, these studies rely on datasets that either exclude robotic actions or treat tactile as a supplementary modality~\cite{jones2025beyond}, limiting their applicability for policy training, confined to perception-only or pick-and-place grasping tasks. The challenges in language-grounded tactile skill learning include: 1) the lack of specialized tactile-action instruction datasets for contact-rich manipulation tasks; and 2) the absence of suitable tactile-language-action models.

\begin{figure}[!t] 
\setlength{\abovecaptionskip}{0cm}
    \centering
\includegraphics[width=0.48\textwidth]{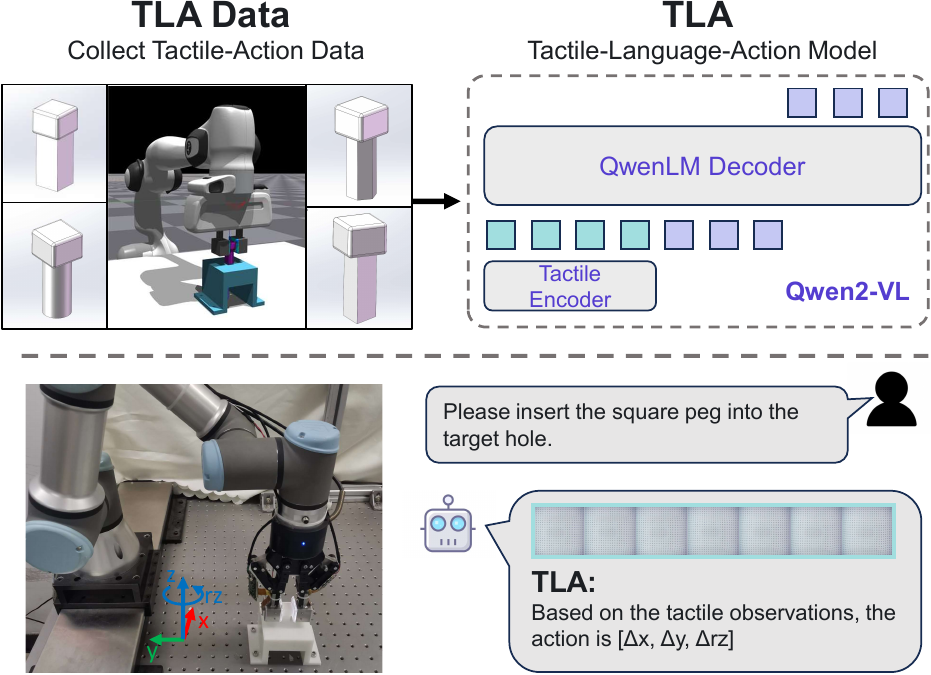}
    \caption{Overview of the proposed Tactile-Language-Action model and tactile-action dataset. The Tactile-Language-Action model is a 7B multimodal language model trained on our tactile-action dataset collected for contact-rich manipulations. With supervised fine-tuning, the TLA can control the robot completing various peg-in-hole assembly tasks. }
\label{fig_overview}
\vspace{-1em}
\end{figure}

To address the aforementioned challenges, we construct a tactile-action instruction dataset tailored for a fingertip tactile peg-in-hole assembly scenario~\cite{chen2024general}. We propose a fine-tuning method for generalist robotic policy learning dubbed the Tactile-Language-Action (TLA) model, demonstrating that cross-modal fine-tuning enables the acquisition of generalized tactile skills through language grounding. The overview of the proposed TLA dataset and model is shown in Fig.~\ref{fig_overview}.

Unlike VLA models, TLA models designed for contact-rich manipulation must effectively handle the sequential tactile information generated by a single contact action. Our key contribution is encoding the generated sequential tactile images into a composite tactile image, which is then processed by an image encoder. This representation, combined with language-grounded reasoning, enables the model to generate robotic actions, allowing our policy to execute interactive contact and collision tasks based on natural language inference. In the fingertip-based peg-in-hole assembly task, our approach can follow natural language operation instructions, such as: “Please insert the square peg into the target hole based on the tactile observation.”

Our results show that TLA outperforms traditional imitation learning methods in terms of single-step motion accuracy and achieves a significant lead in the success rate of actual shaft-hole assembly (multi-steps). Notably, in terms of generalization, TLA demonstrates strong adaptability to varying assembly clearances and peg geometries. Trained solely on assembly data with a 2.0 mm clearance, TLA achieves over 85\% assembly success rates on tasks with 1.6 mm and 1.0 mm clearances. In our experiments, we collected a dataset comprising 24k tactile sequences and corresponding robot action trajectories, generated using a high-fidelity tactile simulator. To the best of our knowledge, the proposed TLA model represents the first language-grounded tactile-action generation framework, marking a significant step toward embodied intelligence.

\section{Related Work}

\subsection{VLM for Robot Manipulation}
Due to the strong perception and reasoning abilities of VLM, RT-2 \cite{brohan2023rt} treats actions in the manipulation dataset as tokens. It fine-tunes on VLM and calls the trained model the Vision-Language-Action (VLA) Model. However, RT-2 is closed-source. RoboFlamingo \cite{li2023vision} and OpenVLA \cite{kim2024openvla}, which follow similar concepts to RT-2, are open-sourced in the robotics community. GR-1 \cite{wu2023unleashing} and GR-2 \cite{cheang2024gr} leverage an Internet-scale video dataset for pre-training. They are then fine-tuned using the manipulation dataset. This method incorporates the physics and dynamics information from the Internet-scale video dataset into the VLA. RDT-1B \cite{liu2024rdt} and PI \cite{black2024pi_0} integrate diffusion concepts into the VLA model framework, demonstrating the effectiveness of training with a diffusion objective. However, the VLA models mentioned above are vision-based and are limited to simple push, pull, and grasp-and-place tasks. This paper explores the effectiveness of using tactile input as an observation for contact-rich peg-in-hole tasks.

\subsection{Tactile-Language Model in Robotics}
Recent studies have attempted to combine tactile information with language models~\cite{yang2024binding}~\cite{fu2024touch}. Fu et al.~\cite{fu2024touch} build a texture-tactile dataset using ChatGPT and propose the Tactile-Vision-Language Model to explore language models' ability in material understanding. However, the dataset scale limits the model's generalization in real-world scenarios. To address this, Cheng et al.~\cite{cheng2024touch100k} propose Touch100k, a large-scale dataset for tactile-vision-based material classification. Unfortunately, these works have not applied tactile information to robot manipulation. Jones et al.~\cite{jones2025beyond} propose FuSe, integrating tactile information into the VLA model to acquire a multimodal robot policy. Different from these works, our work focuses on establishing a connection between tactile perception and actions based on a pre-trained language model. To the best of our knowledge, TLA is the first language-action model solely based on tactile perception.

\begin{figure*}[h] 
\setlength{\abovecaptionskip}{0cm}
    \centering
\includegraphics[width=0.98\textwidth]{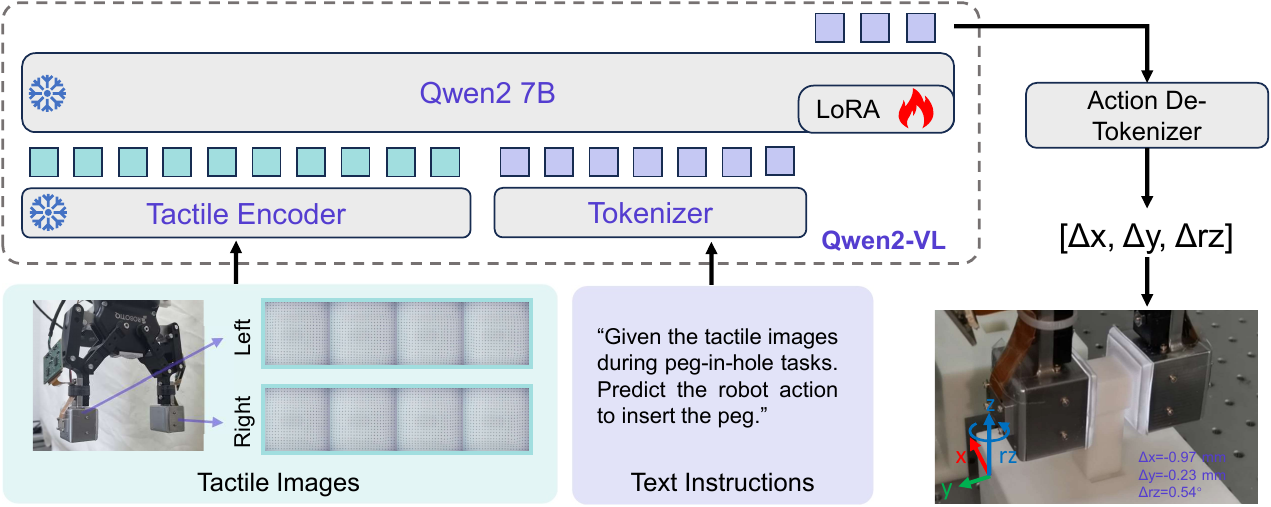}
    \caption{Architecture of Tactile-Language-Action model. Given the tactile images and text instruction, TLA predicts the 3-dimensional robot actions. TLA consists of two modules: (1) Tactile Encoder captures the tactile representations for the language model. (2) The Language Model fuses and understands tactile and text information to predict the robot's actions.}
\label{fig_mtd_1}
\end{figure*}
        
\begin{figure}[h] 
\setlength{\abovecaptionskip}{0cm}
    \centering
\includegraphics[width=0.48\textwidth]{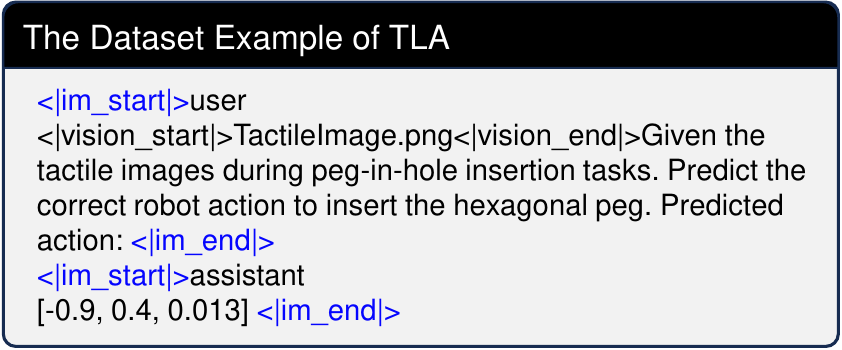}
    \caption{An example of TLA Dataset.}
\label{data_example}
\vspace{-0.5cm}
\end{figure}

\section{Dataset}

In this section, we introduce a tactile-action instruction dataset collected in the process of peg-in-hole assembly. In this task, the robot equipped with GelStereo 2.0 visuotactile sensors~\cite{zhang2023gelstereo} attempts to insert a peg into the corresponding hole based on fingertip tactile sensing and language instruction. For efficient data collection, we build a simulation environment for this task in NVIDIA Isaac Gym. The deformation of the visuotactile sensor during interaction is simulated based on Finite Element Method (FEM) using a Flex physics engine. The tactile imprint rendering method proposed in~\cite{cui2023tactile} is utilized to simulate tactile images. To narrow the sim-real gap, a tactile image obtained from a real sensor is employed for texture mapping instead of a manually designed pattern. In this way, we can obtain high-fidelity tactile images during insertion attempts.

The process of the peg-in-hole task is described as follows. The gripper first grasps a peg with a shape description and moves to the top of the corresponding hole with a random 3-DOFs misalignment in the x-axis, y-axis, and rotation around the z-axis (denoted by $rz$). Then, the gripper moves down to attempt insertion. If a collision occurs between the peg and hole during the downward movement, this attempt is deemed failed, and the gripper lifts up waiting for the next attempt. The tactile image sequence during collision is recorded to infer the robot action $(\Delta x, \Delta y, \Delta rz)$ that adjusts the peg pose. If the collision does not occurred while the gripper moving down to a predetermined position, the task is deemed successful. The maximum number of attempts is 15. Otherwise, the task fails.

In simulation, we employ a random insertion policy for this peg-in-hole task. For each attempt, we save the tactile image sequence and the peg-hole pose error. Then, we create the action labels $(\Delta \hat{x}, \Delta \hat{y}, \Delta \hat{rz})$ from peg-hole pose errors $(e_x,e_y,e_{rz})$.
\begin{equation}
     \Delta \hat{x} = \left\{
    \begin{array}{c}
        \mathbb{F}_{clip}(-e_x+c/2, -\delta, 0),\quad if \  e_x \geq 0, \\
        \mathbb{F}_{clip}(-e_x-c/2, 0, \delta),\quad if \  e_x >0,
    \end{array}
    \right. 
\end{equation}
\begin{equation}
     \Delta \hat{y} = \left\{
    \begin{array}{c}
        \mathbb{F}_{clip}(-e_y+c/2, -\delta, 0),\quad if \  e_y \geq 0, \\
        \mathbb{F}_{clip}(-e_y-c/2, 0, \delta),\quad if \  e_y >0,
    \end{array}
    \right. 
\end{equation}
\begin{equation}
    \Delta \hat{rz} = \mathbb{F}_{clip}(-e_{rz}, -1.5^\circ, 1.5^\circ )
\end{equation}
where $c$ is the assembly clearance. $\mathbb{F}_{clip}$ limits the action to a certain range to improve the stability of the policy. $\delta$ is set to 1 mm according to our experience. In this paper, the dataset is collected only using pegs with 2.0 mm clearance.

To facilitate model training, the collected interaction data is transformed into the instruction format. The \texttt{<|im\_start|>} and \texttt{<|im\_end|>} tokens mark the start and end of each dialogue round. Tactile images are input with \texttt{<|vision\_start|>} and \texttt{<|vision\_end|>}, denoting start and end of visual inputs. A text instruction defines the task, clarifies the peg type, and poses requirements. The robot's actions during data collection are saved as ground-truth. An example of the TLA data is shown in Fig.~\ref{data_example}.

\section{Tactile-Language-Action Model}

This section introduces the proposed Tactile-Language-Action model, as shown in Fig~\ref{fig_mtd_1}. TLA is built based on Qwen2-VL \cite{qwen2-vl}, which includes a Vision Transformer (ViT) \cite{vit} for encoding the visual input and the Qwen2 language model \cite{qwen2} for understanding multimodal information and generating text. This section first gives the details of encoding tactile information with ViT, then introduces how to predict robot actions with the language model.

\subsection{Tactile Encoder}

As mentioned in the previous section, the tactile information of visuotactile sensor is represented in the form of images. During robot manipulation, the tactile images continuously change according to the robotic gripper's contact state. Therefore, the tactile encoder needs to process two temporally aligned image sequences. 

The tactile encoder's challenge lies in extracting temporal variation from input tactile images. To address this, we synthesize two tactile image sequences into a single image, converting temporal information into spatial space to facilitate ViT-based feature extraction. Specifically, the input image set denotes as $I=\{I_l^t,I_r^t;t=0,1,2,3\}$, where $I_l^t$ and $I_r^t$ represent the tactile images of the left and right fingertip at timestamp $t$, respectively. These eight images are arranged in a $3\times3$ grid, with the last grid filled by a white image, and are resized to $616\times616$ as model input.

We use the ViT of Qwen2-VL as the tactile encoder, which is shown in Fig. \ref{fig_mtd_1}. The concatenated tactile images are processed by the tactile encoder to obtain tactile features. Moreover, a Multi-Layer Perceptron (MLP) layer is used to further compress the tactile features within a $2\times2$ range to a single token. Therefore, for the input tactile image $I \in \mathbb{R}^{616 \times 616}$, 1936 tactile tokens can be obtained after passing through the ViT with a patch size of 14.

\subsection{Action Prediction with Language Model}

This section gives the details of using the Qwen2 language model to predict robot actions, as shown in Fig. \ref{fig_mtd_1}. The inputs of the language model are tactile tokens and language tokens, which are acquired by encoding the original multimodal inputs through the tactile encoder and the tokenizer, respectively. The Qwen2 7B is TLA's backbone model and is fine-tuned on the TLA dataset.

The encoding of continuous numbers with a discrete tokenizer affects the model's performance in number-sensitive tasks \cite{momentor}. Previous work \cite{kim2024openvla} simply overwrite the least used tokens as ``special tokens" and assigns bin IDs to each token. Unlike prior approaches, we retain the numerical encoding scheme to ensure that the pre-training acquired numerical knowledge is effectively leveraged in robotic manipulation tasks.

However, the tokenizer of Qwen2 encodes numbers individually. Since there are many decimals in the robot action data, this redundant information increases the difficulty of the model training. To this end, we simplify the ground-truth by scaling all actions with a ratio and rounding to integers. Specifically, the processing procedure is calculated as $A_{gt}=A_{raw} \cdot s$, where $A_{gt}\in \mathbb{R}^3$, $A_{raw}\in \mathbb{R}^3$, $s\in \mathbb{R}^3$ are the ground-truth action, raw action and scale factors. 

\subsection{Training and Inference}

Previous work has shown that the VLMs achieve better performance with the frozen visual encoder during training~\cite{qwen2-vl} \cite{Prismatic}. Therefore, we freeze the parameters of the Tactile Encoder during fine-tuning. Additionally, we use Low-Rank Adaptation (LoRA)~\cite{lora} to efficiently fine-tune the Qwen2 7B language model. The ground-truth actions are used as labels to calculate the Next Token Prediction loss, which is calculated as follows,
\begin{equation}
\mathcal{L}_{\text{NTP}} = - \sum_{t=1}^{T} p(y_t) \log P(y_t \mid y_{<t}, x)
\end{equation}
where $T$ and $y_t$ are the length of the action sequence and the ground-truth token at step $t$, respectively. $y_{<t}$ denotes the previously generated tokens, replaced with ground-truth during training. $x$ represents the input, including tactile image tokens and text instruction tokens.

During inference, TLA sequentially predicts the probability distribution of the robot's actions based on the input tactile observations and instruction texts. The generation process is carried out through beam search until the end token is generated. Finally, the Action-De-Tokenizer maps all the generated probabilities to natural language text according to the vocabulary and converts them into floating-point numbers that can be executed by the robot.

\section{Experiment}

In this section, we investigate the effectiveness of our proposed TLA model for language-conditioned tactile learning in a fingertip peg-in-hole assembly task. Specifically, we focus on using only tactile observations to generate assembly action commands using TLA, enabling language-conditioned tactile skill learning. The goal of our experiment is to answer the following questions:

\begin{itemize}
    \item Compared to conventional imitation learning approaches that train from scratch, does our language-conditioned TLA model achieve a better understanding and modeling of the relationship between tactile observations and task-specific action commands?
    \item Can language models enhance the generalization capability of TLA across different objects and variations of the peg-in-hole assembly task?
    \item Is the trained TLA model capable of effectively controlling the robot to successfully perform a variety of peg-in-hole assembly tasks with different geometries and constraints?
\end{itemize}

\begin{figure*}[h] 
\setlength{\abovecaptionskip}{0cm}
    \centering
\includegraphics[width=0.98\textwidth]{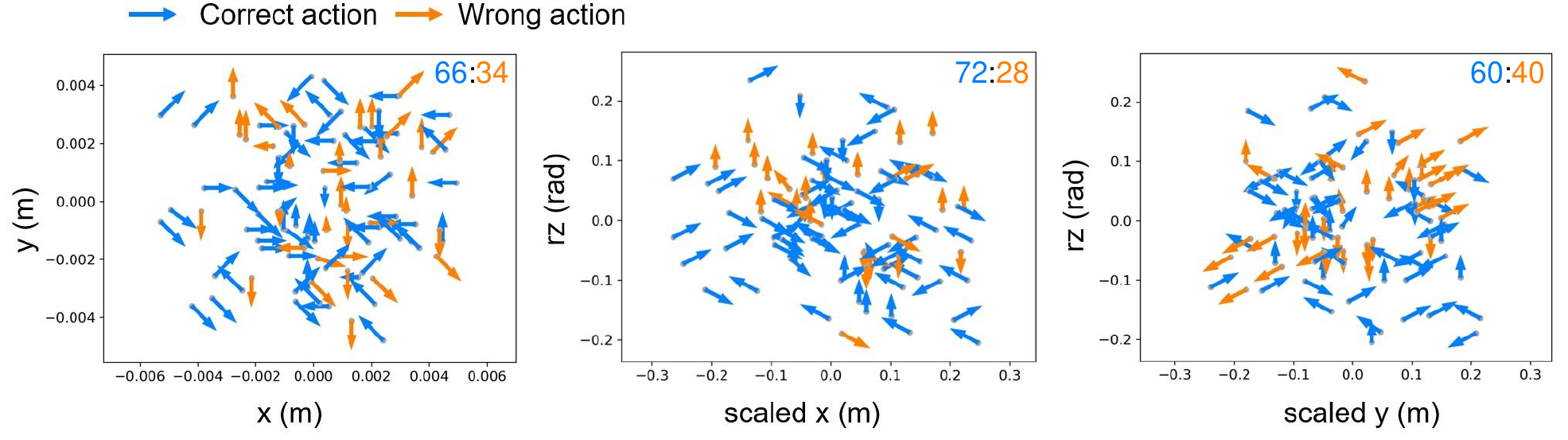}
    \caption{Visualization of action prediction on the single-peg test set. By combining each dimension's actions in pairs, three sub-figures are given, where the correct and wrong actions are marked with different colors. Scaled means the raw data is scaled by a factor of 50 for visualization. The results show that TLA can predict the correct actions for various deviations.}
\label{fig_e1}
\end{figure*}

\subsection{Baseline and Metrics}

To answer the above questions, we compare the following
baseline methods and ablation methods with the proposed TLA on various tasks.
\begin{itemize}
    \item \textbf{Behavior Cloning (BC)}~\cite{torabi2018behavioral}: The ResNet-50~\cite{he2016deep} is employed as the policy network. The network takes the tactile image as input and outputs the robot action, which is trained using a supervised manner.
    \item \textbf{Diffusion Policy (DP)}~\cite{chi2023diffusion}: The diffusion policy in \cite{chi2023diffusion} utilizes the conditional denoising diffusion process to learn the peg-in-hole assembly policy.
    \item \textbf{Single-Peg TLA (SP-TLA)}: The TLA model trained on the square peg insertion dataset. 
    \item \textbf{Multi-Peg TLA (MP-TLA)}: The TLA model trained on the square and triangular peg insertion dataset. 
\end{itemize}

To evaluate different policies, we first evaluate the performance of the model's effective action generation with \textbf{Goal Convergence Rate (GCR)}, which is defined as the percentage of all the output actions that are correct in the $x$, $y$, and $rz$ directions. Then, the L1 distance is used to evaluate the performance of different models, which is calculated as follows,
\begin{equation}
     L_1 = \frac{1}{m} \sum_{i = 1}^{m} |y_i - \hat{y}_i|,
\end{equation}
\noindent where $m$, $y_i$, $\hat{y}_i$ are the sample numbers in GCR, action prediction, and ground-truth, respectively. 
We calculate the L1 distance of each action dimension and present the results.

\begin{table}[!t]
    \setlength{\abovecaptionskip}{0cm}
    \caption{Comparison of different methods on the single-peg test set.}
    \label{tab_e1_2}
    \centering
    \setlength{\tabcolsep}{0.7mm}{
    \begin{tabular}{c|cccc}
    \hline\hline
       \textbf{Method} & \textbf{GCR (\%)}$\boldsymbol{\uparrow}$          & \textbf{L1 x (mm)}$\boldsymbol{\downarrow}$     & \textbf{L1 y (mm)}$\boldsymbol{\downarrow}$    & \textbf{L1 rz (deg)}$\boldsymbol{\downarrow}$\\ \hline
    \textbf{BC~\cite{torabi2018behavioral}} & 10.4          & 0.803          & 0.302          & 0.205            \\
    \textbf{DP~\cite{chi2023diffusion}}  & 8.5          & 0.370          & 0.382          & 0.568            \\
    \textbf{SP-TLA~(Ours)} & \textbf{12.5} & \textbf{0.079} & \textbf{0.122} & \textbf{0.173}  \\ \hline\hline
    \end{tabular}
    }
\end{table}	

\begin{table*}[!h]
    \setlength{\abovecaptionskip}{0cm}
    \caption{Comparison of different methods on the multi-peg test set.}
    \label{tab_e2}
    \centering
    \setlength{\tabcolsep}{1.2mm}{
    \begin{tabular}{c|cccc|cccc}
    \hline\hline
    \multirow{2}{*}{\textbf{Method}} & \multicolumn{4}{c|}{\textbf{ID}}                                           & \multicolumn{4}{c}{\textbf{OOD}}                                           \\
                            & \textbf{GCR(\%)}$\boldsymbol{\uparrow}$         & \textbf{L1 x (mm)}$\boldsymbol{\downarrow}$     & \textbf{L1 y (mm)}$\boldsymbol{\downarrow}$      & \textbf{L1 rz (deg)}$\boldsymbol{\downarrow}$     & \textbf{GCR (\%)}$\boldsymbol{\uparrow}$            & \textbf{L1 x (mm)}$\boldsymbol{\downarrow}$      & \textbf{L1 y (mm)}$\boldsymbol{\downarrow}$      & \textbf{L1 rz (deg)}$\boldsymbol{\downarrow}$      \\ \hline
    \textbf{BC~\cite{torabi2018behavioral}}                      & \textbf{18.4} & 0.260          & 0.655          & 0.186          & 0.152          & 0.286          & 0.722          & 0.246          \\
    \textbf{DP~\cite{chi2023diffusion}}                      & 7.4          & 0.371          & 0.348          & 0.480          & 0.080          & 0.386          & 0.369          & 0.544          \\
    \textbf{MP-TLA~(Ours)}                    & \textbf{18.4} & \textbf{0.102} & \textbf{0.114} & \textbf{0.135} & \textbf{0.165} & \textbf{0.121} & \textbf{0.102} & \textbf{0.184} \\ \hline\hline
    \end{tabular}
    }
\end{table*}	
\subsection{Comparison on Single-Peg Inserting Tasks}

To answer the first question, this subsection compares the TLA with the baseline methods on the TLA dataset.

\textbf{Experimental Setup} We take 8k square peg insertion data from the TLA dataset to compare the performance of TLA and the baseline methods. Specifically, the training set and the test set are split into 6k and 2k, respectively. Each training sample consists of 8 tactile images, which correspond to the left and right fingertips with a time length of 4. The label is the robot action $(\Delta x, \Delta y, \Delta rz)$, which represents the movement in the horizontal plane and the rotation around the z-axis, respectively. The TLA model is trained on 8 Nvidia A6000 GPUs for 20 epochs.

\textbf{Results} The performance of different methods on the single-peg test set is shown in Table \ref{tab_e1_2}. The GCR results indicate that TLA has more correct actions than the baselines. Furthermore, the L1 results show that the correct actions predicted by TLA have more accurate step lengths than others, where the L1 of the x-direction is reduced by 78\% compared to the second-best method. The superior L1 performance implies that TLA is expected to complete tasks with fewer operation steps, demonstrating better manipulation efficiency.

The visualization results of model predictions are shown in Fig.~\ref{fig_e1}, which includes three sub-figures by combining different action dimensions. The starting point of each arrow represents the pose deviation between the current peg and the correct insertion position, while the arrow direction indicates the predicted action movement. The correct and wrong predictions are marked in blue and orange, respectively. For the x-y pair, an action is considered correct if it reduces the position deviation. For the pairs of x-rz and y-rz, an action is correct if it reduces the deviation in at least one dimension. Experimental results demonstrate that TLA's actions perform well in planar translation, effectively guiding the robot toward the target hole. However, in the translation-rotation visualization, more incorrect actions are observed along the y-axis. We attribute this to the limitation of 2D tactile images, which effectively represent tactile information parallel to the gripper (x-axis) but poorly represent information perpendicular to the gripper (y-axis). This imbalanced feature representation increases the difficulty of y-axis predictions.

\subsection{Comparison on Multi-Peg Inserting Tasks}

To address the second question, this subsection trains TLA on various peg types and compares it with baseline methods.

\textbf{Experimental Setup}  
We take 16k samples of square and triangular pegs for training, equally split. An additional 8k samples are collected for evaluation: 4k for square/triangular pegs and 4k for round/hexagonal pegs. The input, output, and metrics align with the single-peg task. The training uses 8 Nvidia A6000 GPUs for 10 epochs.

\textbf{Results} The comparison of different methods on the multi-peg test set is presented in Table \ref{tab_e2}. In the in-distribution (ID) experiments, TLA achieves the lowest L1 error on seen pegs, indicating that the actions predicted by TLA are not only correct but also exhibit more precise and stable step lengths during execution. For the out-of-distribution (OOD) setting, TLA maintains performance that is closely aligned with that on the ID set, highlighting its strong generalization capabilities when facing unseen peg geometries. Notably, TLA-OOD does not suffer from any significant degradation compared to TLA-ID, in stark contrast to traditional imitation learning baselines, which exhibit pronounced performance drops under distribution shifts. This demonstrates that TLA is remarkably robust to variations in peg shapes and is capable of effectively transferring its learned manipulation skills to novel, unseen configurations, substantially outperforming conventional approaches in terms of generalization.

\begin{table}[!t]
    \setlength{\abovecaptionskip}{0cm}
    \caption{Comparison of different methods in insertion tasks with various clearances.}
    \label{tab_e3}
    \centering
    \setlength{\tabcolsep}{0.8mm}{
    \begin{tabular}{c| c c c c c c |c c }
        \hline\hline 
        \multirow{2}{*}{\textbf{Method}} &\multicolumn{2}{c}{\textbf{2.0~mm}} & \multicolumn{2}{c}{\textbf{1.6~mm}} & \multicolumn{2}{c|}{\textbf{1.0~mm}} & \multicolumn{2}{c}{\textbf{Total}} \\
        &\textbf{Suc}$\uparrow$ &\textbf{Step}$\downarrow$ &\textbf{Suc}$\uparrow$ &\textbf{Step}$\downarrow$ & \textbf{Suc}$\uparrow$ &\textbf{Step}$\downarrow$ &\textbf{Suc}$\uparrow$ &\textbf{Step}$\downarrow$ \\
        \hline
        \textbf{BC~\cite{torabi2018behavioral}} & 44 & 2.60 & 32 & 4.00 & 18 & 4.44 & 31 & 3.68 \\
        \textbf{DP~\cite{chi2023diffusion}} & 58& \textbf{2.45} & 43 & 4.35 & 20 & 4.70 &40   &3.83\\
        \textbf{SP-TLA} & \textbf{96}&3.15 &\textbf{94} &3.77&74&4.37&88&3.76\\
        \textbf{MP-TLA}  &94&3.04&86&\textbf{3.30}&\textbf{90}&\textbf{4.35}&\textbf{90}&\textbf{3.56}\\
        \hline\hline 
    \end{tabular}}
    \vspace{-0.2cm}
\end{table}	
\begin{figure*}[t] 
\setlength{\abovecaptionskip}{0cm}
    \centering
\includegraphics[width=0.98\textwidth]{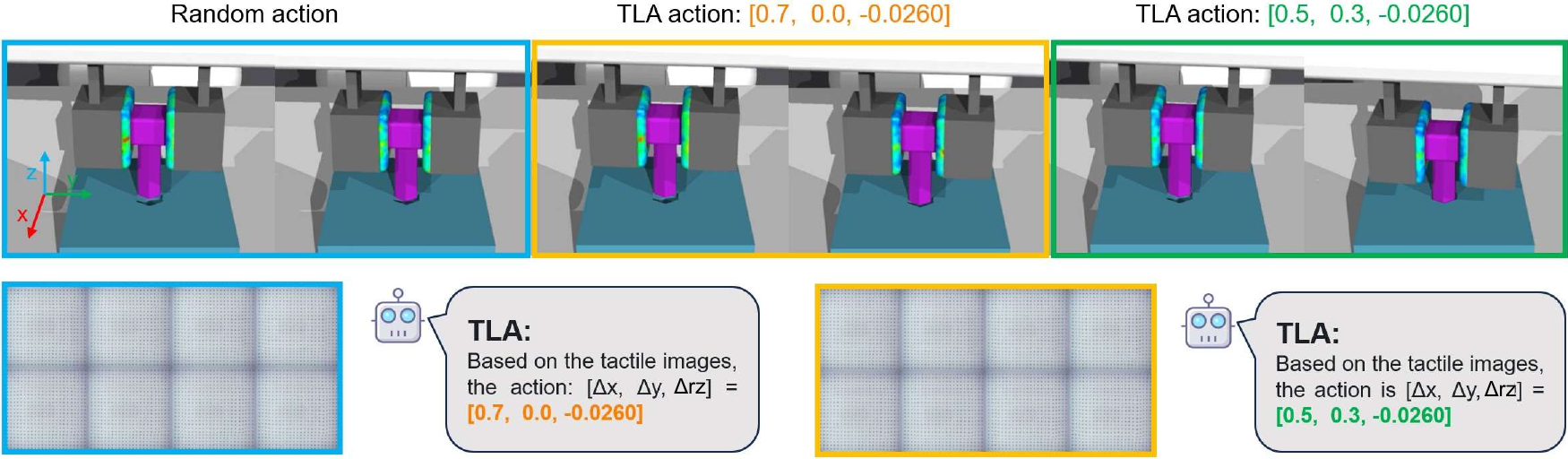}
    \caption{Snapshots of the robot successfully inserting the hexagonal peg into the target hole with the assistance of our proposed TLA.}
\label{fig_exp_e3_success}
\end{figure*}
\subsection{Robotic Insertion Tasks}

In this subsection, the manipulation performance of different models is evaluated in the robotic insertion tasks to answer the third question.

\textbf{Experimental Setup} We test the manipulation performance of different models in two types of peg-in-hole assembly tasks. First, we evaluate the model performance with different assembly clearances, including the square peg-in-hole assembly tasks with assembly clearances of 2.0 mm, 1.6 mm, and 1.0 mm, respectively. Then, we report the performance of different models on square, triangular, round, and hexagonal peg insertion tasks, where all assembly clearances are set to 2.0 mm.
Each task is repeated 50 times to calculate the final results.

\begin{table}[!t]
    \setlength{\abovecaptionskip}{0cm}
    \caption{Comparison of different methods in insertion tasks with various peg types.}
    \label{tab_e4}
    \centering
    \setlength{\tabcolsep}{0.8mm}{
    \begin{tabular}{c| c c c c | c c c c }
        \hline\hline 
        &\multicolumn{4}{c|}{\textbf{ID}} & \multicolumn{4}{c}{\textbf{OOD}} \\
        \textbf{Method} &\multicolumn{2}{c}{\textbf{Square}} & \multicolumn{2}{c|}{\textbf{Triangle}} & \multicolumn{2}{c}{\textbf{Round}} & \multicolumn{2}{c}{\textbf{Hexagon}} \\
        &\textbf{Suc}$\uparrow$ &\textbf{Step}$\downarrow$ &\textbf{Suc}$\uparrow$ &\textbf{Step}$\downarrow$ & \textbf{Suc}$\uparrow$ &\textbf{Step}$\downarrow$ &\textbf{Suc}$\uparrow$ &\textbf{Step}$\downarrow$ \\
        \hline
        \textbf{BC~\cite{torabi2018behavioral}} & 60 & 3.37 & 84 & 2.67 & 62 & 2.84 & 66& \textbf{2.30} \\
        \textbf{DP~\cite{chi2023diffusion}} & 60 & 3.60 & 60& 3.40 & 56 & 2.89 & 54 & 2.37 \\
        \textbf{SP-TLA}  & \textbf{96} & 3.15 &76 &2.82&92&2.80&90&2.78\\
        \textbf{MP-TLA}  &94&\textbf{3.04}&\textbf{92}&\textbf{1.80}&\textbf{94}&\textbf{2.66}&\textbf{96}&3.00\\
        \hline\hline 
    \end{tabular}}
\end{table}	

At the beginning of each episode, the robot has already grasped the peg in its gripper, and the end-effector is randomly set at a starting position near the target hole. Then, the robot attempts the first insertion and obtains tactile images. Subsequently, TLA or other models predict the robot's action based on the tactile observation and control the robot to insert again. The robot continues attempting until the insertion is successful or reaches the maximum attempt number 15. The success rate and the average steps of successful episodes are used to evaluate the performance of all models.

\textbf{Results}  The quantitative comparison results of different methods on various assembly clearances are shown in Table~\ref{tab_e3}. Compared with the baseline methods, TLA has achieved a superior success rate and manipulation efficiency, surpassing the second-best method by 50\% in the success rate. Even with the challenging assembly clearance of 1.0 mm, our TLA model also achieves superior performance than baselines, showing its excellent generalization for assembly clearances. The quantitative comparison results of different methods on various peg types are shown in Table \ref{tab_e4}. Both the two TLA models have achieved a better success rate and steps than baseline methods, demonstrating excellent generalization performance across different peg types. In particular, the experiments on OOD set show that the MP-TLA model performs better than the SP-TLA model. We attribute this to the training data of different pegs can further enhance the generalization of TLA.  

\begin{figure}[t] 
\setlength{\abovecaptionskip}{0cm}
    \centering
\includegraphics[width=0.48\textwidth]{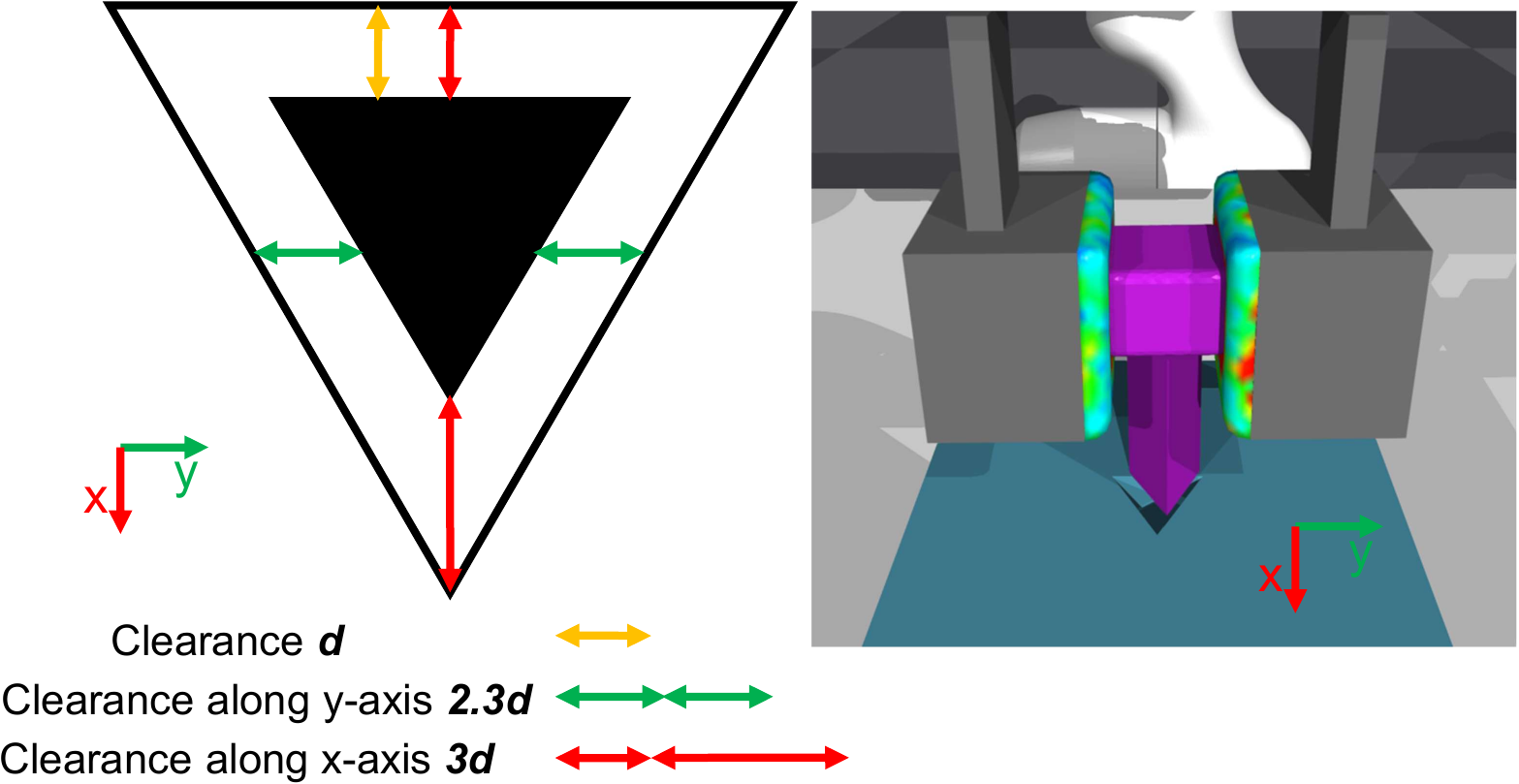}
    \caption{Visualization and analysis of failure case. The TLA fails to insert the triangular peg because the triangular peg has a smaller assembly clearance along the y-axis, and the TLA model based on 2D observations has poor prediction performance along the y-axis. }
\label{fig_exp_e3}
\end{figure}

\textbf{Visualization} The visualization result of TLA controlling the robot to successfully insert the peg is shown in Fig. \ref{fig_exp_e3_success}. First, the robot fails to insert the peg with the initial action. Then, TLA predicts the adjusted robot actions based on the tactile observation during previous attempts. Finally, TLA successfully controls the robot to approach and insert the peg after two rounds of attempts. 

The visualization of a failure case is shown in Fig. \ref{fig_exp_e3}, where TLA fails to control the robot to insert the triangular peg into the hole. We attribute this failure case to the triangular hole having different allowable deviations along the x-axis and the y-axis. Specifically, when assembly clearance is d, the allowable assembly deviation along the x-axis is 3d, while that in the y-axis is only 2.3d. Moreover, since the current tactile information is 2D images, its representation in the direction perpendicular to the gripper (i.e. y-axis) is relatively poor than the other two axes. Therefore, this imbalanced feature representation increases the difficulty for the model to understand the contact information along the y-axis. As a result, due to the model's poor performance in the y-axis direction and the smaller allowable deviation of the triangular shape in the y-axis direction, TLA exhibits poor performance in triangular peg insertion tasks.

\section{Discussion and Limitations}

In this study, we introduce TLA, a Tactile-Language-Action model designed for contact-rich manipulation scenarios. Through a cross-modal fine-tuning process, TLA enables the acquisition of generalized tactile skills via language grounding. We further demonstrate that TLA significantly outperforms traditional imitation learning methods in a challenging fingertip tactile peg-in-hole assembly task, achieving superior assembly success rates. Moreover, it exhibits strong generalization capabilities across variations in assembly clearness and peg shapes.

Despite these promising results, TLA still has several limitations. One notable limitation is that the model does not rigorously capture tactile temporal information. Instead, it encodes temporal cues through spatial arrangements, which may not fully exploit the sequential nature of tactile data. Future work should explore more effective representations and encoding strategies tailored to sequential tactile perception. Another area for improvement lies in the selection of tactile signal formats, which remains relatively rudimentary in this study. Different tactile representations, such as 2D tactile images, 2D contact depth maps, and 3D tactile point clouds, offer unique advantages, and investigating their integration with VLMs for tactile skill learning could further enhance the model’s capabilities. Additionally, the current action de-tokenization process is relatively simplistic, leaving room for refinement. A more sophisticated decoding mechanism could improve the interpretability and precision of learned actions, making it a valuable direction for joint exploration between VLA and TLA models. 

We will deploy TLA in real-world environments to evaluate its sim-to-real generalization, testing how well the policy learned in simulation transfers under real-world uncertainties. All related data, models, and code are available on our project website, where we will also share updates on our real-world experiments. We welcome the community to follow our progress and explore these challenges together.






\bibliographystyle{IEEEtran}
\bibliography{IEEEabrv, paper}

\begin{thebibliography}{10}
\providecommand{\url}[1]{#1}
\csname url@samestyle\endcsname
\providecommand{\newblock}{\relax}
\providecommand{\bibinfo}[2]{#2}
\providecommand{\BIBentrySTDinterwordspacing}{\spaceskip=0pt\relax}
\providecommand{\BIBentryALTinterwordstretchfactor}{4}
\providecommand{\BIBentryALTinterwordspacing}{\spaceskip=\fontdimen2\font plus
\BIBentryALTinterwordstretchfactor\fontdimen3\font minus \fontdimen4\font\relax}
\providecommand{\BIBforeignlanguage}[2]{{%
\expandafter\ifx\csname l@#1\endcsname\relax
\typeout{** WARNING: IEEEtran.bst: No hyphenation pattern has been}%
\typeout{** loaded for the language `#1'. Using the pattern for}%
\typeout{** the default language instead.}%
\else
\language=\csname l@#1\endcsname
\fi
#2}}
\providecommand{\BIBdecl}{\relax}
\BIBdecl
\renewcommand{\BIBentryALTinterwordstretchfactor}{4}

\bibitem{billard2019trends}
A.~Billard and D.~Kragic, ``Trends and challenges in robot manipulation,'' \emph{Science}, vol. 364, no. 6446, p. eaat8414, 2019.

\bibitem{cui2021hand}
S.~Cui, R.~Wang, J.~Hu \emph{et~al.}, ``In-hand object localization using a novel high-resolution visuotactile sensor,'' \emph{IEEE Transactions on Industrial Electronics}, vol.~69, no.~6, pp. 6015--6025, 2021.

\bibitem{cui2021toward}
J.~Cui and J.~Trinkle, ``Toward next-generation learned robot manipulation,'' \emph{Science robotics}, vol.~6, no.~54, p. eabd9461, 2021.

\bibitem{calandra2018more}
R.~Calandra, A.~Owens, D.~Jayaraman \emph{et~al.}, ``More than a feeling: Learning to grasp and regrasp using vision and touch,'' \emph{IEEE Robotics and Automation Letters}, vol.~3, no.~4, pp. 3300--3307, 2018.

\bibitem{li2020review}
Q.~Li, O.~Kroemer, Z.~Su \emph{et~al.}, ``A review of tactile information: Perception and action through touch,'' \emph{IEEE Transactions on Robotics}, vol.~36, no.~6, pp. 1619--1634, 2020.

\bibitem{wang2020swingbot}
C.~Wang, S.~Wang, B.~Romero \emph{et~al.}, ``Swingbot: Learning physical features from in-hand tactile exploration for dynamic swing-up manipulation,'' in \emph{International Conference on Intelligent Robots and Systems (IROS)}, 2020, pp. 5633--5640.

\bibitem{qi2023general}
H.~Qi, B.~Yi, S.~Suresh \emph{et~al.}, ``General in-hand object rotation with vision and touch,'' in \emph{Conference on Robot Learning}, 2023, pp. 2549--2564.

\bibitem{lee2020making}
M.~A. Lee, Y.~Zhu, P.~Zachares \emph{et~al.}, ``Making sense of vision and touch: Learning multimodal representations for contact-rich tasks,'' \emph{IEEE Transactions on Robotics}, vol.~36, no.~3, pp. 582--596, 2020.

\bibitem{dong2021tactile}
S.~Dong, D.~K. Jha, D.~Romeres \emph{et~al.}, ``Tactile-rl for insertion: Generalization to objects of unknown geometry,'' in \emph{IEEE International Conference on Robotics and Automation (ICRA)}, 2021, pp. 6437--6443.

\bibitem{team2024octo}
O.~M. Team, D.~Ghosh, H.~Walke \emph{et~al.}, ``Octo: An open-source generalist robot policy,'' \emph{arXiv preprint arXiv:2405.12213}, 2024.

\bibitem{brohan2023rt}
A.~Brohan, N.~Brown, J.~Carbajal \emph{et~al.}, ``Rt-2: Vision-language-action models transfer web knowledge to robotic control,'' \emph{arXiv preprint arXiv:2307.15818}, 2023.

\bibitem{naveed2023comprehensive}
H.~Naveed, A.~U. Khan, S.~Qiu \emph{et~al.}, ``A comprehensive overview of large language models,'' \emph{arXiv preprint arXiv:2307.06435}, 2023.

\bibitem{kim2024openvla}
M.~J. Kim, K.~Pertsch, S.~Karamcheti \emph{et~al.}, ``Openvla: An open-source vision-language-action model,'' \emph{arXiv preprint arXiv:2406.09246}, 2024.

\bibitem{wen2024tinyvla}
J.~Wen, Y.~Zhu, J.~Li \emph{et~al.}, ``Tinyvla: Towards fast, data-efficient vision-language-action models for robotic manipulation,'' \emph{arXiv preprint arXiv:2409.12514}, 2024.

\bibitem{o2024open}
A.~O’Neill, A.~Rehman, A.~Maddukuri \emph{et~al.}, ``Open x-embodiment: Robotic learning datasets and rt-x models: Open x-embodiment collaboration 0,'' in \emph{IEEE International Conference on Robotics and Automation (ICRA)}, 2024, pp. 6892--6903.

\bibitem{zhou2023language}
H.~Zhou, X.~Yao, Y.~Meng \emph{et~al.}, ``Language-conditioned learning for robotic manipulation: A survey,'' \emph{arXiv preprint arXiv:2312.10807}, 2023.

\bibitem{yang2024binding}
F.~Yang, C.~Feng, Z.~Chen \emph{et~al.}, ``Binding touch to everything: Learning unified multimodal tactile representations,'' in \emph{Proceedings of the IEEE/CVF Conference on Computer Vision and Pattern Recognition}, 2024, pp. 26\,340--26\,353.

\bibitem{cheng2024touch100k}
N.~Cheng, C.~Guan, J.~Gao \emph{et~al.}, ``Touch100k: A large-scale touch-language-vision dataset for touch-centric multimodal representation,'' \emph{arXiv preprint arXiv:2406.03813}, 2024.

\bibitem{fu2024touch}
L.~Fu, G.~Datta, H.~Huang \emph{et~al.}, ``A touch, vision, and language dataset for multimodal alignment,'' \emph{arXiv preprint arXiv:2402.13232}, 2024.

\bibitem{jones2025beyond}
J.~Jones, O.~Mees, C.~Sferrazza \emph{et~al.}, ``Beyond sight: Finetuning generalist robot policies with heterogeneous sensors via language grounding,'' \emph{arXiv preprint arXiv:2501.04693}, 2025.

\bibitem{chen2024general}
W.~Chen, J.~Xu, F.~Xiang \emph{et~al.}, ``General-purpose sim2real protocol for learning contact-rich manipulation with marker-based visuotactile sensors,'' \emph{IEEE Transactions on Robotics}, pp. 1509--1526, 2024.

\bibitem{li2023vision}
X.~Li, M.~Liu, H.~Zhang \emph{et~al.}, ``Vision-language foundation models as effective robot imitators,'' \emph{arXiv preprint arXiv:2311.01378}, 2023.

\bibitem{wu2023unleashing}
H.~Wu, Y.~Jing, C.~Cheang \emph{et~al.}, ``Unleashing large-scale video generative pre-training for visual robot manipulation,'' \emph{arXiv preprint arXiv:2312.13139}, 2023.

\bibitem{cheang2024gr}
C.-L. Cheang, G.~Chen, Y.~Jing \emph{et~al.}, ``Gr-2: A generative video-language-action model with web-scale knowledge for robot manipulation,'' \emph{arXiv preprint arXiv:2410.06158}, 2024.

\bibitem{liu2024rdt}
S.~Liu, L.~Wu, B.~Li \emph{et~al.}, ``Rdt-1b: a diffusion foundation model for bimanual manipulation,'' \emph{arXiv preprint arXiv:2410.07864}, 2024.

\bibitem{black2024pi_0}
K.~Black, N.~Brown, D.~Driess \emph{et~al.}, ``$\backslash pi\_0 $: A vision-language-action flow model for general robot control,'' \emph{arXiv preprint arXiv:2410.24164}, 2024.

\bibitem{zhang2023gelstereo}
C.~Zhang, S.~Cui, S.~Wang \emph{et~al.}, ``Gelstereo 2.0: An improved gelstereo sensor with multimedium refractive stereo calibration,'' \emph{IEEE Transactions on Industrial Electronics}, vol.~71, no.~7, pp. 7452--7462, 2023.

\bibitem{cui2023tactile}
S.~Cui, Y.~Wang, S.~Wang \emph{et~al.}, ``Tactile imprint simulation of gelstereo visuotactile sensors,'' in \emph{IEEE International Conference on Mechatronics and Automation (ICMA)}, 2023, pp. 650--656.

\bibitem{qwen2-vl}
P.~Wang, S.~Bai, S.~Tan \emph{et~al.}, ``Qwen2-vl: Enhancing vision-language model's perception of the world at any resolution,'' \emph{arXiv preprint arXiv:2409.12191}, 2024.

\bibitem{vit}
A.~Dosovitskiy, L.~Beyer, A.~Kolesnikov \emph{et~al.}, ``An image is worth 16x16 words: Transformers for image recognition at scale,'' \emph{arXiv preprint arXiv:2010.11929}, 2020.

\bibitem{qwen2}
A.~Yang, B.~Yang, B.~Hui \emph{et~al.}, ``Qwen2 technical report,'' \emph{arXiv preprint arXiv:2407.10671}, 2024.

\bibitem{momentor}
L.~Qian, J.~Li, Y.~Wu \emph{et~al.}, ``Momentor: Advancing video large language model with fine-grained temporal reasoning,'' \emph{arXiv preprint arXiv:2402.11435}, 2024.

\bibitem{Prismatic}
S.~Karamcheti, S.~Nair, A.~Balakrishna \emph{et~al.}, ``Prismatic vlms: Investigating the design space of visually-conditioned language models,'' in \emph{Forty-first International Conference on Machine Learning}, 2024.

\bibitem{lora}
E.~J. Hu, Y.~Shen, P.~Wallis \emph{et~al.}, ``Lora: Low-rank adaptation of large language models.'' \emph{ICLR}, p.~3, 2022.

\bibitem{torabi2018behavioral}
F.~Torabi, G.~Warnell, and P.~Stone, ``Behavioral cloning from observation,'' \emph{arXiv preprint arXiv:1805.01954}, 2018.

\bibitem{he2016deep}
K.~He, X.~Zhang, S.~Ren \emph{et~al.}, ``Deep residual learning for image recognition,'' in \emph{Proceedings of the IEEE conference on computer vision and pattern recognition}, 2016, pp. 770--778.

\bibitem{chi2023diffusion}
C.~Chi, Z.~Xu, S.~Feng \emph{et~al.}, ``Diffusion policy: Visuomotor policy learning via action diffusion,'' \emph{The International Journal of Robotics Research}, p. 02783649241273668, 2023.

\end{thebibliography}

\end{document}